\documentclass[a4paper]{article}

\usepackage{graphicx}
\usepackage{amsmath,amssymb} 
\usepackage{color}
\usepackage{booktabs}
\usepackage{xspace}
\usepackage{subfig}
\usepackage{overpic}
\usepackage{hyperref}
\usepackage{rotating}
\usepackage{authblk}

\newcommand{\etal}{\emph{et al.}\@\xspace}
\newcommand*{\eg}{e.g.\@\xspace}
\newcommand*{\ie}{i.e.\@\xspace}
\newcommand*{\cf}{c.f.\@\xspace}

\begin{document}

\title{Understanding Regularization to\\ Visualize Convolutional Neural Networks} 



\author[1]{Maximilian Baust*}
\author[1]{Florian Ludwig*}
\author[2]{Christian Rupprecht}
\author[1]{Matthias Kohl}
\author[1]{Stefan Braunewell}
\affil[1]{\small{Konica Minolta Laboratory Europe\\
\texttt{maximilian.baust@konicaminolta.eu}, \texttt{http://research.konicaminolta.eu/}}, *equal contribution}
\affil[2]{Computer Aided Medical Procedures and Augmented Reality (CAMPAR), Technical University of Munich}


\maketitle

\begin{abstract}
Variational methods for revealing visual concepts learned by convolutional neural networks have gained significant attention during the last years.
Being based on noisy gradients obtained via back-propagation such methods require the application of regularization strategies.
We present a mathematical framework unifying previously employed regularization methods.
Within this framework, we propose a novel technique based on Sobolev gradients which can be implemented via convolutions and does not require specialized numerical treatment, such as total variation regularization.
The experiments performed on feature inversion and activation maximization demonstrate the benefit of a unified approach to regularization, such as sharper reconstructions via the proposed Sobolev filters and a better control over reconstructed scales.
\end{abstract}

\section{Introduction}
\label{sec:intro}
One of the great advantages of convolutional neural networks (CNNs) is the fact that they are differentiable.
This property facilitates application of effective gradient-based optimization methods not only for learning representations, but also for visualizing them.
As pointed out by Olah, Mordvintsev and Schubert \cite{olah2017feature}, the latter task requires regularization techniques such as Gaussian filtering.
In this article we study previously proposed regularization strategies for feature visualization from a more general point, derive a general theoretical framework for them and propose novel variants based on this framework.
\subsection{Motivation}
With CNN-based methods penetrating safety-critical domains such a health care or autonomous driving, there is a growing demand for techniques that can visualize representations learned by a network or reveal the reason for a particular decision of it. 
Especially in the area of computer vision, visualization techniques, \eg for exploring trained visual representations, have thus become an important field of research in the last years.
Selvaraju \etal \cite{selvaraju2016grad} distinguished the benefit of such visualization techniques according to the performance that a specific CNN-based method achieves in comparison human raters:
To identify failure in case of inferior or sub-human performance, to establish trust and confidence in case the performance is on par with humans, and to educate the user in case of super-human performance.

As noted by Olah \etal \cite{olah2018the}, most current visualization techniques are not yet sufficient for achieving the aforementioned goals and we are certainly at the very beginning of truly understanding the decisions made by convolutional neural networks.
Nevertheless, it is very likely that the building blocks of future techniques take advantage of the fact that convolutional neural networks are differentiable and it is thus no suprise that many techniques developed for this purpose are variational.
This means that they can be rendered as energy minimization problems, which includes several techniques for computing natural pre-images proposed by Mahendran and Vedaldi \cite{mahendran2016visualizing}.
Following the definition in \cite{mahendran2016visualizing}, natural pre-images are naturally looking images which serve a certain purpose such as maximally stimulating the activation of a certain neuron or class (activation maximization) or yielding the same feature representation as a truly natural image (inversion).
Research on visualization techniques that can be formulated as energy minimization problems and more specifically on their numerical treatment is thus very important, which is the goal of this article.

As noted by Mahendran and Vedaldi \cite{mahendran2016visualizing} as well as Olah, Mordvintsev and Schubert \cite{olah2017feature}, variational visualization techniques greatly benefit from regularization, \eg by incorporating the total variation (TV) as a natural and unbiased prior, because the gradient of the optimized energy usually suffers from a significant amount of noise and high-frequency components.
One of the reasons for this effect are high-frequency patterns arising from strided convolutional layers and pooling layers \cite{olah2017feature}.

\begin{figure}
\centering
\begin{minipage}{1.0\textwidth}
\begin{overpic}[clip,trim=0cm 0cm 0cm 0cm,width=0.32\textwidth]{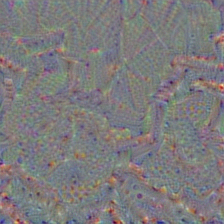}
\put(5,5){\textcolor{white}{Sobolev fluid}}
\end{overpic}
\hfill
\begin{overpic}[clip,trim=0cm 0cm 0cm 0cm,width=0.32\textwidth]{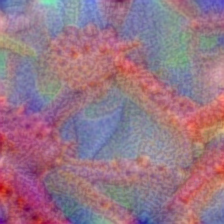}
\put(5,5){\textcolor{white}{Sobolev elastic}}
\end{overpic}
\hfill
\begin{overpic}[clip,trim=0cm 0cm 0cm 0cm,width=0.32\textwidth]{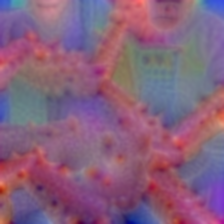}
\put(5,5){\textcolor{white}{Sobolev fluid-elastic}}
\end{overpic}
\end{minipage}\\[0.5em]
\begin{minipage}{1.0\textwidth}
\begin{overpic}[clip,trim=0cm 0cm 0cm 0cm,width=0.32\textwidth]{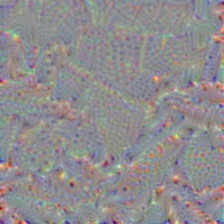}
\put(5,5){\textcolor{white}{Gaussian fluid}}
\end{overpic}
\hfill
\begin{overpic}[clip,trim=0cm 0cm 0cm 0cm,width=0.32\textwidth]{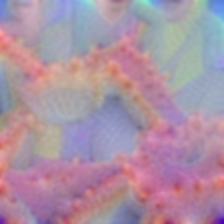}
\put(5,5){\textcolor{white}{Gaussian elastic}}
\end{overpic}
\hfill
\begin{overpic}[clip,trim=0cm 0cm 0cm 0cm,width=0.32\textwidth]{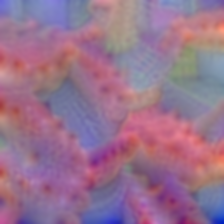}
\put(5,5){\textcolor{white}{Gaussian fluid-elastic}}
\end{overpic}
\end{minipage}
\begin{minipage}{1.0\textwidth}
\begin{overpic}[clip,trim=0cm 0cm 0cm 0cm,width=0.32\textwidth]{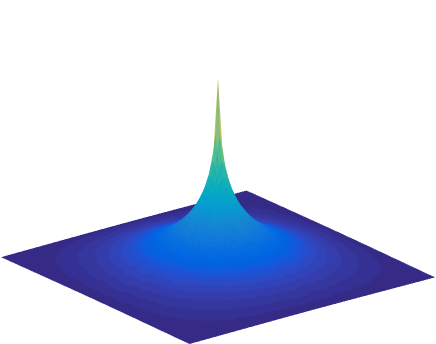}
\put(5,70){\textcolor{black}{Sobolev kernel}}
\end{overpic}
\hfill
\begin{overpic}[clip,trim=0cm 0cm 0cm 0cm,width=0.32\textwidth]{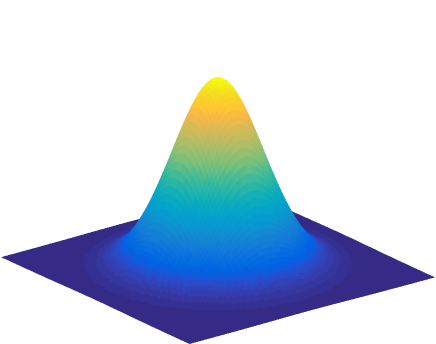}
\put(5,70){\textcolor{black}{Gaussian kernel}}
\end{overpic}
\hfill
\begin{overpic}[clip,trim=0cm 0cm 0cm 0cm,width=0.32\textwidth]{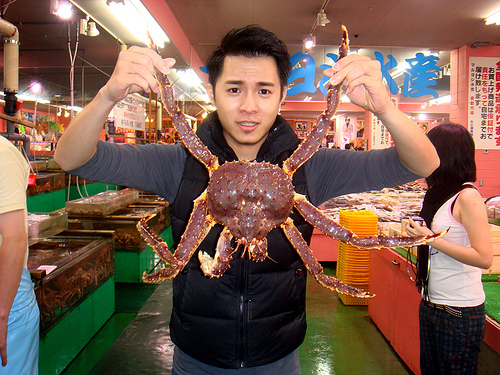}
\put(5,5){\textcolor{white}{exemplary image}}
\end{overpic}
\end{minipage}
\caption{\textbf{Qualitative Comparison of Various Demons Optimization Schemes and Filters:} Activation maximization for the \textit{king crab} class in the VGG19 network pre-trained on ImageNet. The Sobolev filter proposed in this work (first row) produces crisper details in comparison to the widely used Gaussian filter (second row). While fluid demons yield results with only fine details, elastic and fluid-elastic demons yield pre-images with both coarse and fine details. Interestingly the latter pre-images also contain class specific biases in ImageNet, \ie people in the background holding the crab (see also the exemplary image).}
\label{fig:SobolevGaussian}
\end{figure}

Taking inspirations from variational image registration, more specifically demons-type approaches \cite{thirion1998image,vercauteren2009diffeomorphic}, and Sobolev gradient methods for vision and image processing \cite{neuberger2009sobolev,chefd2002flows,sundaramoorthi2007sobolev,calder2010image} we propose a unified view on first-order gradient-based optimization schemes for computing pre-images and offer a powerful alternative to TV regularization.
We demonstrate in the remainder of this article that the combination of demons-type optimization schemes and Sobolev-type regularization yield convincing results for variational visualization techniques such as activation maximization and feature inversion, which becomes already visible in Fig.~\ref{fig:SobolevGaussian}.
\subsection{Related Work}
\label{sec:RelWork}
During the last decade, a plethora of methods for understanding neural networks has been developed \cite{selvaraju2016grad,maaten2008visualizing,erhan2009visualizing,simonyan2013deep,zeiler2014visualizing,springenberg2014striving,nguyen2015deep,nguyen2016plug,fong2017interpretable,kindermans2017patternnet,kindermans2017reliability,sundararajan2017axiomatic,olah2017feature,olah2018the}.
As the focus of this work is on optimization and regularization techniques, we refer the interested reader to the recent articles of G\"{u}n \etal \cite{grun2016taxonomy} of Olah \etal \cite{olah2018the}.
In the latter one, the authors characterize these methods nicely according to their actual goal, \ie visualization, attribution and dimensionality reduction, and present a unified approach for feature visualization based on these concepts.

Focusing on regularization techniques for variational methods to computing pre-images, related work has employed various strategies: no regularization \cite{erhan2009visualizing,szegedy2013intriguing}, total variation regularization \cite{mahendran2016visualizing}, blurring the pre-image or the update computed in every iteration with Gaussian filters \cite{nguyen2015deep,oygard2015glasses,mordvintsev2015inceptionism}, bilateral filters \cite{tyka2016art}, jitter and multiscale representations \cite{mordvintsev2016deepdream} or trained and untrained convolutional networks and auto-encoders \cite{dosovitskiy2015inverting,nguyen2016synthesizing,nguyen2016plug,ulyanov2017practical}.

Besides yielding impressive results, filtering-based methods have the advantage that they are both easy to implement and lead to unconditionally stable optimization schemes.
However, Gaussian filtering is not edge-preserving and might sometimes result in too blurry images.
To avoid such effects, total variation regularization and bilateral filtering are viable alternatives, but require special numerical treatment (\ie splitting schemes or step size restrictions in case of total variation) or the adjustment of contrast parameters (in case of bilateral filtering).
By utilizing Sobolev-filters for regularization, \cf. Sec.~\ref{sec:methods} we are able to provide the best of these two worlds: less over-smoothing the Gaussian filtering and an efficient and stable update scheme that is easy to implement.
\subsection{Summary of Contributions}
In Sec.~\ref{sec:methods}, we review Gaussian filtering from a theoretical point of view, derive a novel class of filters, \ie Sobolev filters, based on the concept of Sobolev gradient methods, and derive a unified framework based on Demons-type optimization schemes.
In Sec.~\ref{sec:experiments} we compare Sobolev filtering to both Gaussian filtering and total variation regularization and demonstrate that the proposed regularization approach yields convincing visual and quantitative results for the tasks of activation maximization and inversion.
Finally, we discuss the relationship of the derived framework to other approaches in Sec.~\ref{sec:discussion} and show that most of the aforementioned works can be classified in a unified way.

\section{Methodology}
\label{sec:methods}
In this section, we will review variational models for computing pre-images from a general point of view (Sec.~\ref{subsec:variation}) as well as Gaussian filtering and its connection to regularization, cf. Sec.~\ref{subsec:gauss_solution} and Sec.~\ref{subsec:gauss_update}.
Next, we derive Sobolev filters based on the concept of gradient flows in Sobolev spaces in Sec.~\ref{subsec:sobolev_gradient} before we propose a general approach for filtering-based regularization in Sec.~\ref{subsec:unified}.
\subsection{Variational Models for Pre-images}
\label{subsec:variation}
Let us consider the general form of a variational model for computing pre-images following the notation of Mahendran and Vedaldi \cite{mahendran2016visualizing}:
\begin{equation}
    \min_{u\in\mathcal{F}}D(\Phi(u),\Phi_0) + \lambda R(u),
    \label{eqn:varProb}
\end{equation}
where $u:\Omega\subset\mathbb{R}^2\rightarrow\mathbb{R}$ is the pre-image to be computed and $\mathcal{F}$ denotes the function space in which $u$ is supposed to lie; for instance, it could be the space of square integrable functions
\begin{equation}
    L^2(\Omega) = \left\{ u:\left\| u\right\|_{L^2} < \infty \right\},\quad\text{where}\quad \left\| u\right\|_{L^2} = \int_\Omega  |u(x)|\; dx.
\end{equation}
$D$ is a data term that measures the proximity of a certain representation $\Phi(u)$ to a reference code $\Phi_0$, $R$ is a regularization term that only depends on $u$, and $\lambda>0$ is a parameter to control the trade-off between data fidelity and regularity.
Let us be more specific and consider more concrete examples:
With $u_0$ being a target image and $\Phi_0 = \Phi(u_0) \in\mathbb{R}^d$ being the target code the task of inversion (or feature reconstruction) can be formulated as
\begin{equation}
    D(\Phi(u),\Phi_0) = \frac{1}{Z}\left\| \Phi(u) - \Phi_0 \right\|^p_2,
    \label{eqn:reconstr}
\end{equation}
where $p=1,2$ and $Z>0$ is a suitable normalization constant.
Choosing $\Phi_0$ to be a $d$-dimensional unit vector $e_i\in\mathbb{R}$ the goal of activation maximization can be achieved by minimizing
\begin{equation}
    D(\Phi(u),\Phi_0) = \frac{1}{Z}\left\langle \Phi(u), e_i \right\rangle,
    \label{eqn:act_max}
\end{equation}
where $Z>0$ is again an appropriate normalization constant.
Examples for $R$ will be discussed in the remainder of this section.

Regardless, of the specific optimization task, however, a gradient descent scheme for \eqref{eqn:varProb} can be written as follows:
\begin{equation}
    u^{t+\tau} = u^t - \tau (\nabla_u D^t +\lambda \nabla_u R^t),
    \label{eqn:grad_desc_plain}
\end{equation}
where $u^t$ denotes the solution at the discrete time step $t$, $\nabla_u D^t = \nabla_u D(\Phi(u^t),\Phi_0)$ denotes the gradient of the data term and $\nabla_u R^t = \nabla_u R(u^t)$ the gradient of the regularization term with respect to $u$.
In the next sections, we will see how \eqref{eqn:grad_desc_plain} varies on case of Gaussian filtering as a regularization strategy.

\subsection{Gaussian Filtering of the Solution}
\label{subsec:gauss_solution}
A widely used strategy for regularizing the computation of pre-images, is to filter the updated solution with a Gaussian:
\begin{equation}
    u^{t+\tau} = G_\sigma \ast (u^t - \tau \nabla_u D^t),
    \label{eqn:elastic}
\end{equation}
where $G_\sigma$ is a Gaussian kernel with standard deviation $\sigma$.
Such an update scheme is called \textit{elastic demons} in the deformable registration community \cite{thirion1998image,vercauteren2009diffeomorphic} as it is equivalent to choosing 
\begin{equation}
    R(u) = \int_\Omega \left\| \nabla u(x) \right\|^2_2\;dx.
    \label{eqn:elastic_reg}
\end{equation}
In other words, filtering the solution $u$ with a Gaussian after applying the update is just a convenient -- and numerically stable -- way of using \eqref{eqn:elastic_reg} as a regularizer.

\subsection{Gaussian Filtering of the Update}
\label{subsec:gauss_update}
A possible regularization strategy comprises in regularizing only the gradient of the data term, \ie
\begin{equation}
    u^{t+\tau} = u^t - \tau G_\sigma \ast \nabla_u D^t,
    \label{eqn:fluid}
\end{equation}
which is commonly known as \textit{fluid demons} update scheme pioneered by Christensen, Rabbitt and Miller \cite{christensen1996deformable} for deformable image registration. 
The underlying idea relates to enforcing an elastic regularization not for the deformation field itself, but for its associated velocity \cite{fischer2004unified,vercauteren2009diffeomorphic}.
Fluid-type update schemes typically lead to less smooth solutions which can be seen by the fact that each update is only regularized once:
\begin{align}
    u^t & = u^0 - \tau G_\sigma \ast \nabla_u D^\tau - \tau G_\sigma \ast \nabla_u D^{2\tau} \ldots - \tau G_\sigma \ast \nabla_u D^{t-\tau}\\
     &= u^0 - \tau G_\sigma \ast \sum^n_{i=1} \nabla_u D^{i\tau},
\end{align}
where $t=n\tau$.
In contrast to this, one can observe an exponential behavior in case of elastic type regularization \eqref{eqn:elastic}:
\begin{align}
    u^t & = G_\sigma \ast ( G_\sigma \ast \ldots G_\sigma \ast (G_\sigma \ast (u^0 - \tau \nabla_u D^\tau) - \tau \nabla_u D^{2\tau}) \ldots - \tau \nabla_u D^{t-\tau} ) \\
     &\approx  G_{n\sigma} \ast u^0 - \tau \sum^n_{i=1} G_{(n-i)\sigma} \ast \nabla_u D^{i\tau},
\end{align}
where we have used the fact that $G_{2\sigma}\approx G_\sigma \ast G_\sigma$ holds in practice\footnote{In the continuous domain, the equality holds, but not in a discretized setting.}.
We will see the fundamental difference between these two schemes in Sec.~\ref{sec:experiments} as it yields a significant difference with respect to the scales present in the computed pre-images.

\subsection{Sobolev Gradients and Filters}
\label{subsec:sobolev_gradient}
It has been demonstrated that Gaussian filtering yields impressive results for the computation of pre-images \cite{oygard2015glasses,nguyen2015deep,mordvintsev2015inceptionism}.
While being easy to implement Gaussian filtering is also prone to cause over-smoothing which is why more edge-preserving regularization strategies have been proposed \cite{mahendran2016visualizing,tyka2016art}.
As noted by Olah \etal \cite{olah2017feature}, changing the metric for gradient computation can have a significant impact on the optimization outcome and this is exactly what we propose in order to obtain a filtering technique that introduces less over-smoothing.

We start by taking a closer look at how the gradient $\nabla_uD^t$ is computed.
From a formal perspective, we compute the first variation of $D$ at $u^t$ and obtain $\nabla_uD^t$ by separating it from an arbitrary test function $v$:
\begin{equation}
   \displaystyle \partial_s \left. D(\Phi(u^t+sv),\Phi_0)\right|_{s=0}=\ldots=\int_\Omega \nabla_uD^t(x) v(x)\; dx\stackrel{!}{=}0.
\end{equation}
The main observation is now that $\nabla D^t$ is computed with respect to the $L^2$ inner product, \ie
\begin{equation}
   \int_\Omega \nabla_uD^t(x) v(x)\; dx = \left\langle \nabla_uD^t , v\right\rangle_{L^2}.
\end{equation}
As this inner product has no intrinsic notion of regularity, gradients computed via this inner product do not have a notion of regularity either.
Sobolev spaces, such as $H^1$ have shown to be a powerful means for obtaining more regular gradients as they are endowed with an inner product which facilitates to measure regularity:
\begin{equation}
   \left\langle v,w\right\rangle_{H^1} = \left\langle v,w\right\rangle_{L_2} + \gamma \left\langle \nabla v,\nabla w\right\rangle_{L_2},
\end{equation}
where $\nabla$ denotes the gradient with respect to $x$.
Due to the fact that the inner product of $H^1$ can be expressed via the inner product of $L_2$ it is possible to compute gradients computed with respect to $\left\langle \cdot,\cdot \right\rangle_{H^1}$ directly from the $L_2$ gradient $\nabla_u D^t$ \cite{neuberger2009sobolev,sundaramoorthi2007sobolev}:
\begin{equation}
    (Id-\gamma \Delta)^{-1} \nabla_u D^t,
    \label{eqn:sobolev_gradient}
\end{equation}
where $Id$ denotes the identity operator and $\Delta$ denotes the Laplacian with respect to $x$.
As $(Id-\gamma \Delta)^{-1}$ does not depend on $u$ it can be implemented via a convolution \cite{calder2010image}, where the resulting Sobolev filter $S_\gamma$ can be obtained by solving the equation system $(Id-\gamma \Delta)^{-1}\delta_0$.
In this context, $\delta_0$ denotes the discretized Dirac impulse at $0$.

To sum up, a Sobolev gradient flow for only optimizing the data term reads
\begin{equation}
    u^{t+\tau} = u^t - \tau S_\gamma \ast \nabla_u D^t.
    \label{eqn:sobolev_flow}
\end{equation}
As a consequence it is similar to a fluid demons scheme, but with a significantly less smooth kernel; it is well known that the convolution with a Gaussian causes the result to be infitely many times differentiable whereas elements of the Sobolev space $H^1$ are only weakly differentiable.

\subsection{A Unified Approach Based on Demons}
\label{subsec:unified}
In Sec.~\ref{subsec:gauss_solution}, \ref{subsec:gauss_update} and \ref{subsec:sobolev_gradient} we have seen how three entirely different regularization approaches yield very similar optimization schemes.
This observation has been made in the deformable registration community \cite{chefd2002flows,vercauteren2009diffeomorphic} leading to the concept of fluid-elastic demons.
Following this concept, we propose a general update scheme for computing pre-images:
\begin{equation}
    u^{t+\tau} = K_e\ast(u^t - \tau K_f \ast \nabla_u D^t),
    \label{eqn:general_scheme}
\end{equation}
where $K_e$ is the elastic filtering kernel and $K_f$ is the fluid filtering kernel.
$K_e$ and $K_f$ could be any kind of smoothing kernel, but in this work we restrict ourselves to Gaussian and Sobolev kernels.
As the total variation is a very popular and powerful regularization strategy, we will review it in Sec.~\ref{subsec:total_variation}, before we continue with the experiments in Sec.~\ref{sec:experiments}.
Moreover, as there are plenty connections to other regularization strategy mentioned in Sec.~\ref{sec:RelWork}, we will discuss them briefly in Sec.~\ref{sec:discussion}.

\subsection{Total Variation Regularization}
\label{subsec:total_variation}
Since its introduction for image restoration by Rudin, Osher and Fatemi \cite{rudin1992nonlinear} in 1992, total variation regularization has been one of the most popular regularizers in image processing and computer vision.
For a modern introduction, we refer the interested reader to the review article by Chambolle \etal \cite{chambolle2010introduction}.
For a differentiable function $u$, the total variation can be written as 
\begin{equation}
    TV(u)=\int_\Omega \left\| \nabla u(x) \right\|_2\;dx.
\end{equation}
Despite its success, total variation minimization is a bit more involved from a mathematical point of view \cite{chambolle2010introduction}.
The reason for this fact can be seen by computing the derivative of $TV(u)$ assuming sufficiently differentiable functions with non-vanishing gradient:
\begin{equation}
    \nabla_u TV(u)(x) = -\text{div}\left( \frac{\nabla u(x)}{\left\| \nabla u(x) \right\|}\right),
\end{equation}
where $\text{div}$ denotes the divergence operator.
This expression is not defined for $\left\| \nabla u(x) \right\|=0$ and thus either numerical relaxations, \ie replacing $\left\| \nabla u(x) \right\|$ by $\left\| \nabla u(x) \right\|+\epsilon$, or sophisticated primal-dual schemes have to be used \cite{chambolle2010introduction}.

\subsection{Implementation Details}
\label{subsec:implementation}
For our experiments we implemented the scheme in \eqref{eqn:general_scheme} using keras 2.1.3 and tensorflow 1.4.0 backend (official tensorflow docker with Ubuntu 16.04 LTS).
All considered network architectures and weights are taken from the official keras implementation\footnote{\url{https://github.com/keras-team/keras}}.
We conducted all experiments on a dedicated workstation with Intel i7-6850K processor, 64GB RAM and two NVIDIA Geforce GTX 1080 Ti graphics cards.
In case of activation maximization, we also investigated the usage of octaves and jitter regularization as implemented by \cite{oygard2015glasses}.
More precisely we used the scales 1.0, 1.1 and 1.2 for defining the octaves and random cropping of the scaled images with a randomly chosen translation magnitude of up to 30\% (w.~r.~t.~input image size) in both directions.
The number of used iterations, chosen step sizes and filter kernel sizes is given reported in Sec.~\ref{sec:experiments}.
In order to keep the comparison of Gaussian filtering and Sobolev filtering comparable, we always adjusted their parameters ($\sigma$ and $\gamma$) such that the support of both filters optimally fits the respective filter size: all filter entries larger than 0.0001 are within a margin on one pixel from the filter boundary.
The code for the proposed, flexible visualization framework will be made publicly available on acceptance.
\section{Experiments}
\label{sec:experiments}
We performed a series of experiments for the tasks of activation maximization and feature inversion to demonstrate the differences of the various regularization strategies.
More precisely we focus on fluid demons schemes, elastic demons schemes and fluid-elastic demons schemes with both Sobolev filters and Gaussian filters.
Moreover, we compared these schemes to total variation regularization.
\subsection{Activation Maximization}
We start by activation maximization, see \eqref{eqn:act_max} in Sec.~\ref{subsec:variation}, and compare various demons schemes, \ie fluid demons, elastic demons, and fluid-elastic demons, for Sobolev filters and Gaussian filters in Fig.~\ref{fig:SobolevGaussian}.
We used the \texttt{VGG19} architecture \cite{simonyan2013deep} and optimized for the \textit{king crab} class of ImageNet, where we only applied slight jitter regularization \cite{mordvintsev2015inceptionism} of two pixels and ran all scheme without octaves, in order to avoid a superposition of regularization effects. For computing the pre-images, $160$ steps gradient descent with a step size of $5$ were used.
The filter size was set to $11\times11$ pixels.

\begin{figure*}[p]
\def\thisfigwidth{0.31\textwidth}
	\begin{minipage}[t]{1\textwidth}
            \begingroup
			\hfill
			\begin{minipage}[c]{0.04\textwidth}
				\begin{turn}{90}\small{VGG19}\end{turn}\hfill
			\end{minipage}%
			\begin{minipage}[c]{0.96\textwidth}
				\begin{overpic}[clip,trim=0cm 0cm 0cm 0cm,width=0.24\textwidth]{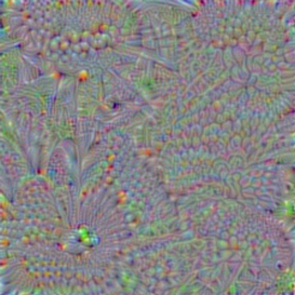}
                \put(5,5){\textcolor{white}{TV}}
                \end{overpic}
                \hfill
                \begin{overpic}[clip,trim=0cm 0cm 0cm 0cm,width=0.24\textwidth]{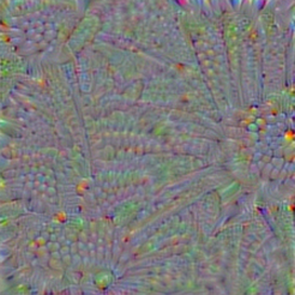}
                \put(5,5){\textcolor{white}{fluid}}
                \end{overpic}
                \hfill
                \begin{overpic}[clip,trim=0cm 0cm 0cm 0cm,width=0.24\textwidth]{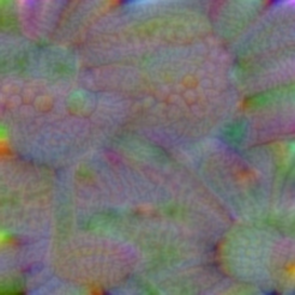}
                \put(5,5){\textcolor{white}{elastic}}
                \end{overpic}
                \hfill
                \begin{overpic}[clip,trim=0cm 0cm 0cm 0cm,width=0.24\textwidth]{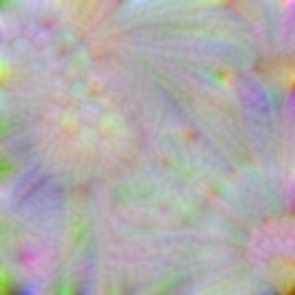}
                \put(5,5){\textcolor{white}{fluid-elastic}}
                \end{overpic}
                \\[0.2em]
                \begin{overpic}[clip,trim=0cm 0cm 0cm 0cm,width=0.24\textwidth]{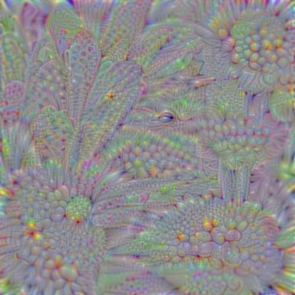}
                \put(5,5){\textcolor{white}{+ octaves \& jitter}}
                \end{overpic}
                \hfill
                \begin{overpic}[clip,trim=0cm 0cm 0cm 0cm,width=0.24\textwidth]{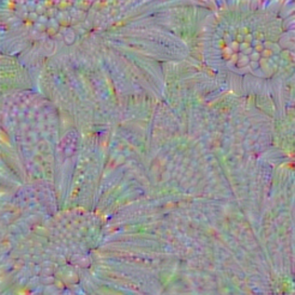}
                \put(5,5){\textcolor{white}{+ octaves \& jitter}}
                \end{overpic}
                \hfill
                \begin{overpic}[clip,trim=0cm 0cm 0cm 0cm,width=0.24\textwidth]{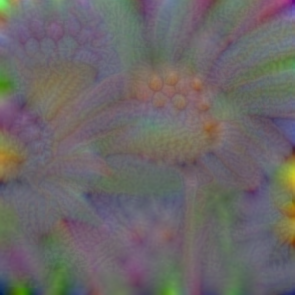}
                \put(5,5){\textcolor{white}{+ octaves \& jitter}}
                \end{overpic}
                \hfill
                \begin{overpic}[clip,trim=0cm 0cm 0cm 0cm,width=0.24\textwidth]{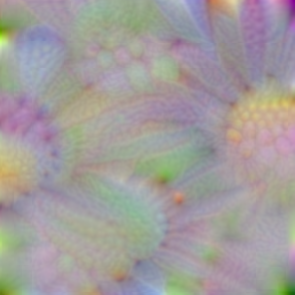}
                \put(5,5){\textcolor{white}{+ octaves \& jitter}}
                \end{overpic}					
			\end{minipage}%
			\par\endgroup
			\vspace{2ex}
			\begingroup
			\hfill
			\begin{minipage}[c]{0.04\textwidth}
				\begin{turn}{90}\small{ResNet50}\end{turn}
			\end{minipage}%
				\begin{minipage}[c]{0.96\textwidth}
				\begin{overpic}[clip,trim=0cm 0cm 0cm 0cm,width=0.24\textwidth]{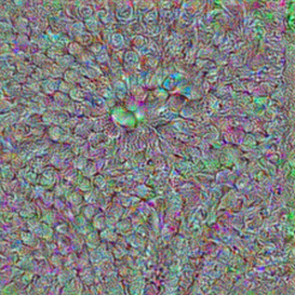}
                \put(5,5){\textcolor{white}{TV}}
                \end{overpic}
                \hfill
                \begin{overpic}[clip,trim=0cm 0cm 0cm 0cm,width=0.24\textwidth]{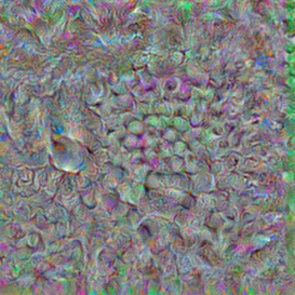}
                \put(5,5){\textcolor{white}{fluid}}
                \end{overpic}
                \hfill
                \begin{overpic}[clip,trim=0cm 0cm 0cm 0cm,width=0.24\textwidth]{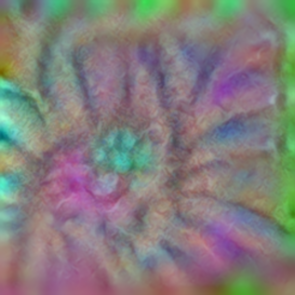}
                \put(5,5){\textcolor{white}{elastic}}
                \end{overpic}
                \hfill
                \begin{overpic}[clip,trim=0cm 0cm 0cm 0cm,width=0.24\textwidth]{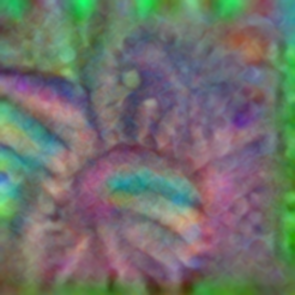}
                \put(5,5){\textcolor{white}{fluid-elastic}}
                \end{overpic}
                \\[0.2em]
                \begin{overpic}[clip,trim=0cm 0cm 0cm 0cm,width=0.24\textwidth]{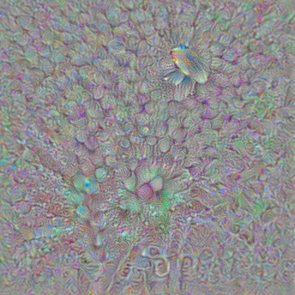}
                \put(5,5){\textcolor{white}{+ octaves \& jitter}}
                \end{overpic}
                \hfill
                \begin{overpic}[clip,trim=0cm 0cm 0cm 0cm,width=0.24\textwidth]{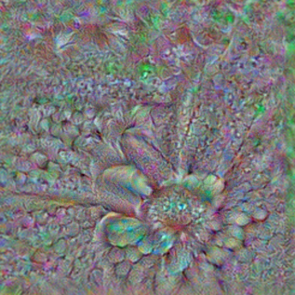}
                \put(5,5){\textcolor{white}{+ octaves \& jitter}}
                \end{overpic}
                \hfill
                \begin{overpic}[clip,trim=0cm 0cm 0cm 0cm,width=0.24\textwidth]{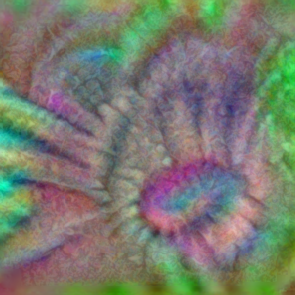}
                \put(5,5){\textcolor{white}{+ octaves \& jitter}}
                \end{overpic}
                \hfill
                \begin{overpic}[clip,trim=0cm 0cm 0cm 0cm,width=0.24\textwidth]{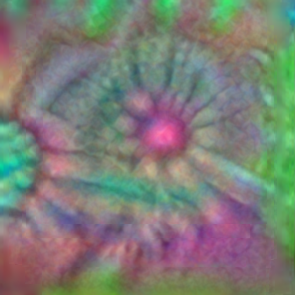}
                \put(5,5){\textcolor{white}{+ octaves \& jitter}}
                \end{overpic}					
			\end{minipage}%
			\par\endgroup
						\vspace{2ex}
		    \begingroup
		    \hfill
		    \begin{minipage}[c]{0.04\textwidth}
		    	\begin{turn}{90}\small{DenseNet121}\end{turn}
		    \end{minipage}%
				\begin{minipage}[c]{0.96\textwidth}
				\begin{overpic}[clip,trim=0cm 0cm 0cm 0cm,width=0.24\textwidth]{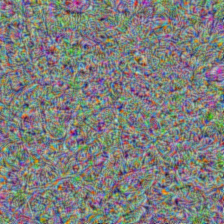}
                \put(5,5){\textcolor{white}{TV}}
                \end{overpic}
                \hfill
                \begin{overpic}[clip,trim=0cm 0cm 0cm 0cm,width=0.24\textwidth]{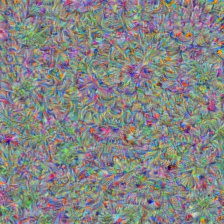}
                \put(5,5){\textcolor{white}{fluid}}
                \end{overpic}
                \hfill
                \begin{overpic}[clip,trim=0cm 0cm 0cm 0cm,width=0.24\textwidth]{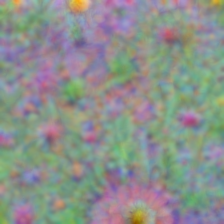}
                \put(5,5){\textcolor{white}{elastic}}
                \end{overpic}
                \hfill
                \begin{overpic}[clip,trim=0cm 0cm 0cm 0cm,width=0.24\textwidth]{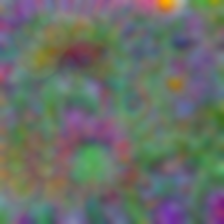}
                \put(5,5){\textcolor{white}{fluid-elastic}}
                \end{overpic}
                \\[0.2em]
                \begin{overpic}[clip,trim=0cm 0cm 0cm 0cm,width=0.24\textwidth]{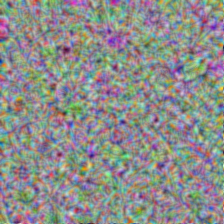}
                \put(5,5){\textcolor{white}{+ octaves \& jitter}}
                \end{overpic}
                \hfill
                \begin{overpic}[clip,trim=0cm 0cm 0cm 0cm,width=0.24\textwidth]{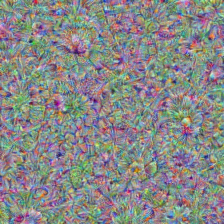}
                \put(5,5){\textcolor{white}{+ octaves \& jitter}}
                \end{overpic}
                \hfill
                \begin{overpic}[clip,trim=0cm 0cm 0cm 0cm,width=0.24\textwidth]{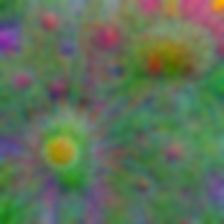}
                \put(5,5){\textcolor{white}{+ octaves \& jitter}}
                \end{overpic}
                \hfill
                \begin{overpic}[clip,trim=0cm 0cm 0cm 0cm,width=0.24\textwidth]{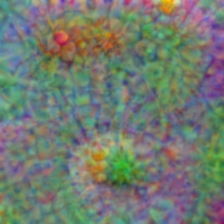}
                \put(5,5){\textcolor{white}{+ octaves \& jitter}}
                \end{overpic}					
			\end{minipage}%
		    \par\endgroup
	\end{minipage}
	\caption{\textbf{Comparison of Demons Schemes for Activation Maximization and Various Architectures on the Daisy Class:} Elastic and fluid-elastic schemes with Sobolev filters produce images containing more scales than fluid schemes (with Sobolev filters) or TV regularized approaches independently of the architecture.}
	\label{fig:act_max_architectures}
\end{figure*}

In Fig.~\ref{fig:act_max_architectures}, we conducted similar experiment for various architectures, \ie \texttt{VGG19}, \texttt{ResNet50}, and \texttt{DenseNet121} \cite{simonyan2013deep,he2016deep,huang2017densely}, for the \textit{daisy} class of ImageNet.
We conducted the experiments both with and without octaves and jitter regularization as described in Sec.~\ref{subsec:implementation}. For various network architectures a different total number of steps and different step sizes were applied, but the Sobolev filter size was kept fixed ($9\times9$ pixels): 

For the \texttt{VGG19} architecture, a total number of $160$ steps were used. In case of the application of multiple octaves, the steps were divided into $100$ steps for the first octave with scale $1.0$, $50$ steps for the second octave with a scale of $1.1$ and $10$ steps for the third octave with a scale of $1.2$. The step size for iterations in the first octave was $5$, for the second octave $2$ and for the third octave $1$.  

For the \texttt{ResNet50} architecture, a total number of $236$ steps were used. In case of the usage of multiple octaves, the steps were divided into 160 steps for the first octave with scale $1.0$, $60$ steps for the second octave with a scale of $1.1$ and $16$ steps for the third octave with a scale of $1.2$. The step size for iterations in the first octave was $60$, for the second octave $20$ and for the third octave $5$. 

For the \texttt{DenseNet121} architecture, a total number of $168$ steps were used. In case of the usage of multiple octaves, the steps were divided into $100$ steps for the first octave with scale $1.0$, $60$ steps for the second octave with a scale of $1.1$ and $8$ steps for the third octave with a scale of $1.2$. The step size for iterations in the first octave was $5$, for the second octave $0.05$ and for the third octave $0.02$. 
\subsection{Reconstruction}
%
To quantitatively assess the quality of the reconstructions, see \eqref{eqn:reconstr} in Sec.~\ref{subsec:variation}, we performed the following experiment:
Given an input image, we computed its reconstruction from a layer inside the network.
To understand if this reconstruction captures the essential part of the original image we fed this image to another architecture trained for the same task.
If the reconstruction is classified correctly, we can assume that visualization was good enough to retain the important information in the image.
For this experiment we use $300$ iterations with a step size of $\tau = 20$ and a filter size of $11\times11$. 
In Table \ref{tab:reconstruction} we show the results of this experiment with \texttt{VGG19} and \texttt{DenseNet} on the ImageNet classification task.
\begin{figure*}[t]%
\centering
    \begin{overpic}[clip,trim=0cm 1cm 0cm 0cm,height=0.21\textwidth]{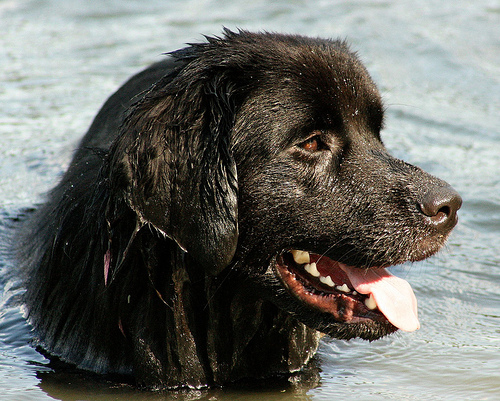}
    \end{overpic}
    \hfill
    \begin{overpic}[clip,trim=0cm 0cm 0cm 0cm,height=0.21\textwidth]{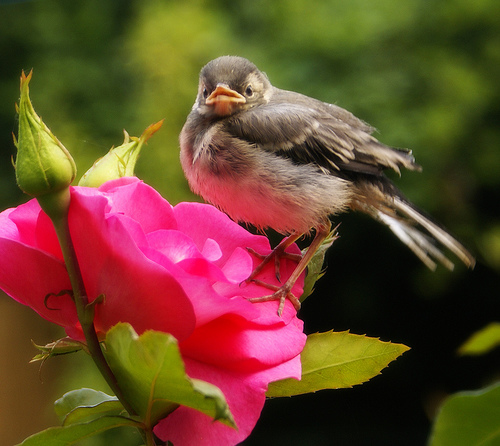}
    \end{overpic}
	\hfill
	\begin{overpic}[clip,trim=0cm 0cm 0cm 0cm,height=0.21\textwidth]{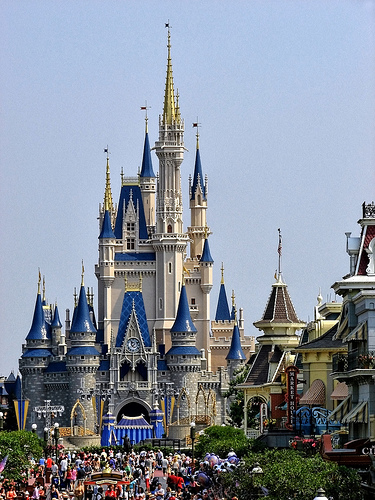}
    \end{overpic}
    \hfill
    \begin{overpic}[clip,trim=0cm 0cm 0cm 0cm,height=0.21\textwidth]{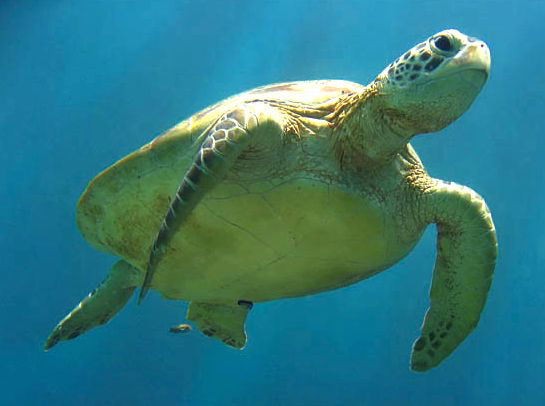}
    \end{overpic}
\caption{\textbf{Test Images Used for Feature Inversion}}
\label{fig:orig_images}
\end{figure*} 
For each class we randomly sampled an image and perform a reconstruction using three different methods on one architecture.
The reconstruction is then classified by the other architecture that was was not used for reconstruction and vice versa.
We further investigated the reconstruction performance of various regularization schemes qualitatively using the images shown in Fig.~\ref{fig:orig_images}; the results of these experiments are shown in Fig.~\ref{fig:reconstruction}.
These experiments were conducted with 2000 steps, a step size of $1.0$ and a Sobolev filter size of $11\times11$ pixels.
All reconstruction experiments were performed without the usage of octaves and jitter.
\begin{table}[h]
    \centering
    \begin{tabular}{lcccc}
    \toprule
          & \multicolumn{2}{c}{\texttt{VGG19} rec. $\rightarrow$ \texttt{DenseNet121}} \hspace{1em} & \multicolumn{2}{c}{\texttt{DenseNet121} rec. $\rightarrow$ \texttt{VGG19}} \\
         \textbf{regularization method}& \textbf{top-1} & \textbf{top-5} & \textbf{top-1} & \textbf{top-5} \\
         \midrule
         \small{total variation}             & \textbf{44.0\%} & 65.7\% &  5.9\% & 14.4\% \\ \midrule
         \small{fluid demons}  & 43.5\% & \textbf{67.3\%} &  4.5\% & 10.4\% \\ \midrule
         \small{fluid-elastic demons} & 13.8\% & 30.6\% & \textbf{10.4\%} & \textbf{21.6\%} \\
         \bottomrule
    \end{tabular}
    \caption{\textbf{Quantitative Reconstruction Assessment:} Average classification score of 1000 images (randomly selected from each class of ImageNet) reconstructed from \texttt{block5\_conv1} of \texttt{VGG19} and \texttt{conv5\_block12\_0\_bn} of \texttt{DenseNet121}, as classified by the other network respectively. We conducted this experiment for total variation regularization, fluidd emons with Sobolev filters and fluid-elastic demons with Sobolev filters.}
    \label{tab:reconstruction}
\end{table}
\begin{figure}%
\centering
    \begingroup
	    \hfill
		\begin{minipage}[c]{0.04\textwidth}
			\begin{turn}{90}\tiny{total variation}\end{turn}\hfill
			\begin{turn}{90}\tiny{\texttt{conv3\_block3\_1\_bn}}\end{turn}
		\end{minipage}
		\begin{minipage}[c]{0.01\textwidth}
		\end{minipage}%
		\begin{minipage}[c]{0.95\textwidth}
			\begin{overpic}[clip,trim=0cm 0cm 0cm 0cm,width=0.24\textwidth]{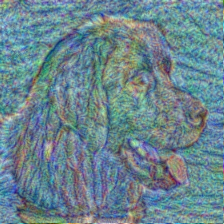}
            \end{overpic}
            \hfill
            \begin{overpic}[clip,trim=0cm 0cm 0cm 0cm,width=0.24\textwidth]{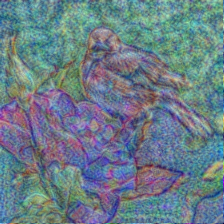}
            \end{overpic}
            \begin{overpic}[clip,trim=0cm 0cm 0cm 0cm,width=0.24\textwidth]{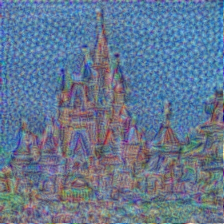}
            \end{overpic}
            \hfill
            \begin{overpic}[clip,trim=0cm 0cm 0cm 0cm,width=0.24\textwidth]{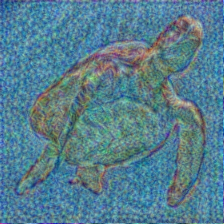}
            \end{overpic}
		\end{minipage}%
	\par\endgroup
    \begingroup
	    \hfill
		\begin{minipage}[c]{0.04\textwidth}
			\begin{turn}{90}\tiny{total variation}\end{turn}\hfill
			\begin{turn}{90}\tiny{\texttt{conv5\_block12\_0\_bn}}\end{turn}
		\end{minipage}
		\begin{minipage}[c]{0.01\textwidth}
		\end{minipage}%
		\begin{minipage}[c]{0.95\textwidth}
			\begin{overpic}[clip,trim=0cm 0cm 0cm 0cm,width=0.24\textwidth]{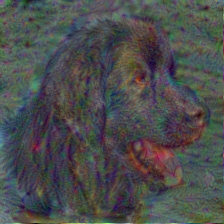}
            \end{overpic}
            \hfill
            \begin{overpic}[clip,trim=0cm 0cm 0cm 0cm,width=0.24\textwidth]{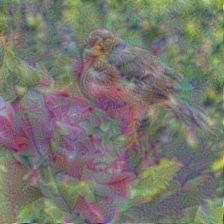}
            \end{overpic}
            \begin{overpic}[clip,trim=0cm 0cm 0cm 0cm,width=0.24\textwidth]{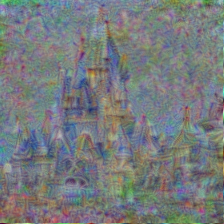}
            \end{overpic}
            \hfill
            \begin{overpic}[clip,trim=0cm 0cm 0cm 0cm,width=0.24\textwidth]{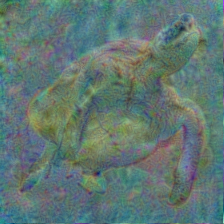}
            \end{overpic}
		\end{minipage}%
	\par\endgroup
    \begingroup
	    \hfill
		\begin{minipage}[c]{0.04\textwidth}
			\begin{turn}{90}\tiny{fluid demons}\end{turn}\hfill
			\begin{turn}{90}\tiny{\texttt{conv3\_block3\_1\_bn}}\end{turn}
		\end{minipage}
		\begin{minipage}[c]{0.01\textwidth}
		\end{minipage}%
		\begin{minipage}[c]{0.95\textwidth}
			\begin{overpic}[clip,trim=0cm 0cm 0cm 0cm,width=0.24\textwidth]{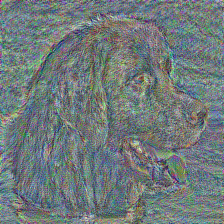}
            \end{overpic}
            \hfill
            \begin{overpic}[clip,trim=0cm 0cm 0cm 0cm,width=0.24\textwidth]{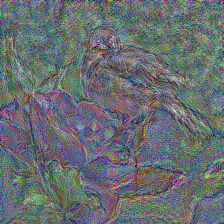}
            \end{overpic}
            \begin{overpic}[clip,trim=0cm 0cm 0cm 0cm,width=0.24\textwidth]{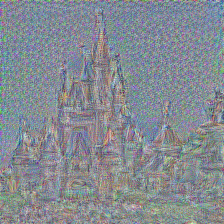}
            \end{overpic}
            \hfill
            \begin{overpic}[clip,trim=0cm 0cm 0cm 0cm,width=0.24\textwidth]{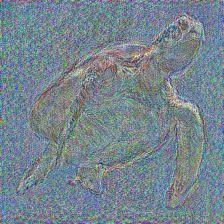}
            \end{overpic}
		\end{minipage}%
	\par\endgroup
	    \begingroup
	    \hfill
		\begin{minipage}[c]{0.04\textwidth}
			\begin{turn}{90}\tiny{fluid demons}\end{turn}\hfill
			\begin{turn}{90}\tiny{\texttt{conv5\_block12\_0\_bn}}\end{turn}
		\end{minipage}
		\begin{minipage}[c]{0.01\textwidth}
		\end{minipage}%
		\begin{minipage}[c]{0.95\textwidth}
			\begin{overpic}[clip,trim=0cm 0cm 0cm 0cm,width=0.24\textwidth]{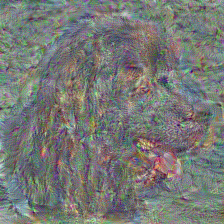}
            \end{overpic}
            \hfill
            \begin{overpic}[clip,trim=0cm 0cm 0cm 0cm,width=0.24\textwidth]{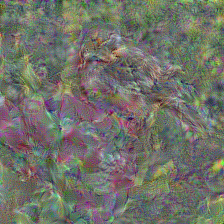}
            \end{overpic}
            \begin{overpic}[clip,trim=0cm 0cm 0cm 0cm,width=0.24\textwidth]{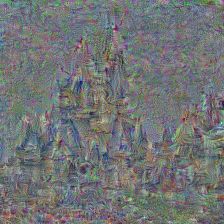}
            \end{overpic}
            \hfill
            \begin{overpic}[clip,trim=0cm 0cm 0cm 0cm,width=0.24\textwidth]{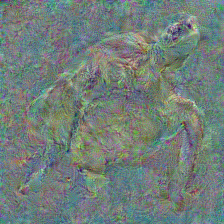}
            \end{overpic}
		\end{minipage}%
	\par\endgroup
	\begingroup
	    \hfill
		\begin{minipage}[c]{0.04\textwidth}
			\begin{turn}{90}\tiny{fluid elastic demons}\end{turn}\hfill
			\begin{turn}{90}\tiny{\texttt{conv3\_block3\_1\_bn}}\end{turn}
		\end{minipage}
		\begin{minipage}[c]{0.01\textwidth}
		\end{minipage}%
		\begin{minipage}[c]{0.95\textwidth}
			\begin{overpic}[clip,trim=0cm 0cm 0cm 0cm,width=0.24\textwidth]{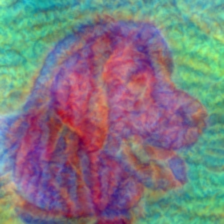}
            \end{overpic}
            \hfill
            \begin{overpic}[clip,trim=0cm 0cm 0cm 0cm,width=0.24\textwidth]{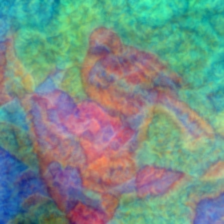}
            \end{overpic}
            \begin{overpic}[clip,trim=0cm 0cm 0cm 0cm,width=0.24\textwidth]{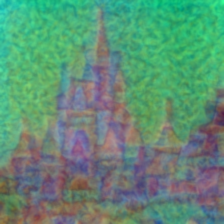}
            \end{overpic}
            \hfill
            \begin{overpic}[clip,trim=0cm 0cm 0cm 0cm,width=0.24\textwidth]{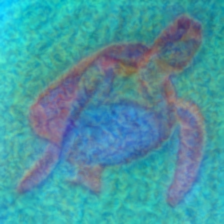}
            \end{overpic}
		\end{minipage}%
	\par\endgroup
	    \begingroup
	    \hfill
		\begin{minipage}[c]{0.04\textwidth}
			\begin{turn}{90}\tiny{fluid elastic demons}\end{turn}\hfill
			\begin{turn}{90}\tiny{\texttt{conv5\_block12\_0\_bn}}\end{turn}
		\end{minipage}
		\begin{minipage}[c]{0.01\textwidth}
		\end{minipage}%
		\begin{minipage}[c]{0.95\textwidth}
			\begin{overpic}[clip,trim=0cm 0cm 0cm 0cm,width=0.24\textwidth]{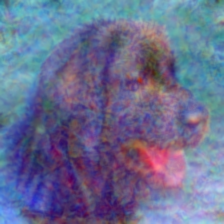}
            \end{overpic}
            \hfill
            \begin{overpic}[clip,trim=0cm 0cm 0cm 0cm,width=0.24\textwidth]{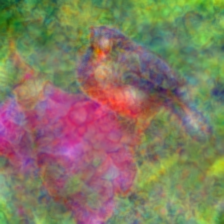}
            \end{overpic}
            \begin{overpic}[clip,trim=0cm 0cm 0cm 0cm,width=0.24\textwidth]{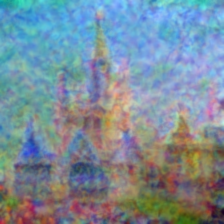}
            \end{overpic}
            \hfill
            \begin{overpic}[clip,trim=0cm 0cm 0cm 0cm,width=0.24\textwidth]{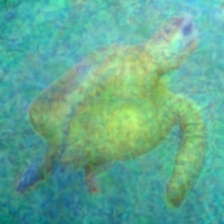}
            \end{overpic}
		\end{minipage}%
	\par\endgroup
\caption{\textbf{Comparison of Various Regularization Techniques for Various Input Images:} We reconstruct the input image from the \texttt{conv3\_block3\_1\_bn} and \texttt{conv5\_block12\_0\_bn} layers of \texttt{DenseNet121} using TV regularization, fluid demons with Sobolev filter and fluid-elastic demons with with Sobolev filters.}
\label{fig:reconstruction}
\end{figure} 

\section{Discussion}
\label{sec:discussion}
We split this section into the discussion of the results and the discussion of the investigated regularization strategies in comparison to the related work cited in Sec.\ref{sec:RelWork}.
\subsection{Discussion of Results}
The experiment in Fig.~\ref{fig:SobolevGaussian} reveal that all demons schemes based on the Sobolev filter yields sharper pre-images and finer details than the schemes using a Gaussian filter for the task of activation maximization, which is no surprise due to the different regularity of both filters.
Moreover, we can observe the generation of both fine and coarse structures for elastic and fluid-elastic schemes.

Similar observations can be made for the comparison of different architectures in Fig.~\ref{fig:act_max_architectures} for the same task, where we observe a better distribution of scales in the final results for the elastic and fluid eleastic schemes.
Interestingly, additional regularization via octaves and jitter does not yield significant improvements.
The reason for this behavior could be the exponential regularization behavior of the elastic scheme discussed in Sec.~\ref{subsec:gauss_solution}.
Furthermore, we can observe that the more recent architectures \texttt{ResNet50} and \texttt{DenseNet121} become slightly harder to visualize with the traditional approaches, which is indicated by slightly less crisp and naturally looking pre-images.
This hints at more complex activation patterns and thus gradients, that need more sophisticated regularization in order to converge to a meaningful result.

Regarding the task of feature inversion (reconstruction), we can observe different levels of smoothness in the reconstruction, \cf Fig.~\ref{fig:reconstruction} .
Fluid demons produces many high frequency features and thus the reconstructions are leas clear for the deeper \texttt{conv5} block. TV regularization is able to keep some structure even for deeper layers while fluid-elastic demons based on Sobolev filters produces the smoothest, most natural looking reconstructions.
This also manifests in Tab.~\ref{tab:reconstruction} where TV and fluid demons work well for the relatively shallow VGG but do not perform well on DenseNet. 

\subsection{The Bigger Picture}
Before concluding this article, we wish to discuss the derived schemes in a broader context.
We have seen that applying a Gaussian filter to the entire solution is equivalent to adding a regularizer to the energy, \cf Sec.~\ref{subsec:gauss_solution}, which is conceptually similar to using the total variation as a regularizer.
On the other hand, regularizing only the gradient of the energy with a Sobolev filter corresponds -- roughly speaking -- to changing the underlying function space or at least its metric, see Sec.~\ref{subsec:sobolev_gradient}.
From a more general point of view, one has two possibilities for regularization: One can either add a regularization term to the respective energy, or modify the underlying function space.
We term the first approach \textit{explicit regularization} and the latter one \textit{implicit regularization}.
This way, it is possible to categorize almost all related works on regularization -- except for jitter regularization -- which have been mentioned in Sec.\ref{sec:RelWork} in Tab.~\ref{tab:relWork}.
It is worth noting that we categorized bilateral filtering \cite{tyka2016art} as explicit regularization strategy due to it similarity to anisotropic diffusion filtering \cite{perona1990scale}.
Furthermore, we considered multiscale representations \cite{mordvintsev2016deepdream} or trained and untrained convolutional networks and auto-encoders \cite{dosovitskiy2015inverting,nguyen2016synthesizing,nguyen2016plug,ulyanov2017practical} as implicit regularization strategies as the lead to parametric function spaces.
Last but not least, it deserves to be mentioned that Kuka{\v{c}}ka, Golkov and Cremers recently released a very nice overview on regularization approaches for training deep networks in general \cite{kukavcka2017regularization}.
As a consequence, all these regularization strategies do also have an influence on the the computation of pre-images via the network itself, but we consider such investigations as future work.

\begin{table}[t]
    \centering
    \begin{tabular}{p{6cm}|p{6cm}}
    \toprule
         \textbf{explicit regularization} & \textbf{implicit regularization} \\ \midrule
         \textbf{total variation} $TV(u)$ \cite{mahendran2016visualizing} & \textbf{fluid demons with Gaussian} \cite{nguyen2015deep} \\ \midrule
         \textbf{elastic} $\int \left\| \nabla u\right\|^2_2$ \cite{oygard2015glasses,mordvintsev2015inceptionism}  a.k.a. \newline
         \textbf{elastic demons with Gaussian} & multiscale representation \cite{mordvintsev2016deepdream} \newline (actually wavelet-type parametrization) \\ \midrule
         bilateral filter \cite{tyka2016art}\newline (non-linear dependency on $u$) & function space parametrized by trained or untrained CNN \cite{dosovitskiy2015inverting,nguyen2016synthesizing,nguyen2016plug,ulyanov2017practical} \\ \midrule
           \textbf{fluid demons with Sobolev filter} & \textbf{elastic and fluid-elastic}\newline \textbf{demons with Sobolev filter} \\
         \bottomrule
    \end{tabular}
    \caption{\textbf{Explicit and Implicit Regularization Strategies:} Approaches in bold are discussed in this article.}
    \label{tab:relWork}
\end{table}

\section{Conclusion}
\label{sec:conclusion}
We have presented a unified categorization of regularization methods for computing pre-images via variational methods.
Based on this categorization, we are able to characterize regularization techniques that have been proposed so far, \cf \ref{tab:relWork}.
Taking inspiration from demons-based deformable image registration and Sobolev gradient methods, we were able to identify novel and previously unexplored optimization and regularization schemes such as fluid-elastic demons schemes including Sobolev filters.
The experiments conducted for activation maximization and feature inversion demonstrate that the derived schemes endow the user with the possibility of better selecting the scale of the reconstruction.
Plain fluid demons schemes with Sobolev filters yield comparable results to total variation regularization and are thus a convenient alternative as convolution-based schemes are both easy to implement and numerically stable.
In addition to this, elastic and fluid-elastic schemes yield a good distribution of coarse and fine scales for the tasks of activation maximization and feature inversion (reconstruction).
Future work might include higher order Sobolev filters or combinations of several regularization schemes.
%

\bibliographystyle{splncs}
\bibliography{references}

\begin{thebibliography}{10}

\bibitem{olah2017feature}
Olah, C., Mordvintsev, A., Schubert, L.:
\newblock Feature visualization.
\newblock Distill (2017) https://distill.pub/2017/feature-visualization.

\bibitem{selvaraju2016grad}
Selvaraju, R.R., Cogswell, M., Das, A., Vedantam, R., Parikh, D., Batra, D.:
\newblock Grad-cam: Visual explanations from deep networks via gradient-based
  localization.
\newblock See https://arxiv. org/abs/1610.02391 v3 \textbf{7}(8) (2016)

\bibitem{olah2018the}
Olah, C., Satyanarayan, A., Johnson, I., Carter, S., Schubert, L., Ye, K.,
  Mordvintsev, A.:
\newblock The building blocks of interpretability.
\newblock Distill (2018) https://distill.pub/2018/building-blocks.

\bibitem{mahendran2016visualizing}
Mahendran, A., Vedaldi, A.:
\newblock Visualizing deep convolutional neural networks using natural
  pre-images.
\newblock International Journal of Computer Vision \textbf{120}(3) (2016)
  233--255

\bibitem{thirion1998image}
Thirion, J.P.:
\newblock Image matching as a diffusion process: an analogy with maxwell's
  demons.
\newblock Medical image analysis \textbf{2}(3) (1998)  243--260

\bibitem{vercauteren2009diffeomorphic}
Vercauteren, T., Pennec, X., Perchant, A., Ayache, N.:
\newblock Diffeomorphic demons: Efficient non-parametric image registration.
\newblock NeuroImage \textbf{45}(1) (2009)  S61--S72

\bibitem{neuberger2009sobolev}
Neuberger, J.:
\newblock Sobolev gradients and differential equations.
\newblock Springer Science \& Business Media (2009)

\bibitem{chefd2002flows}
Chefd'Hotel, C., Hermosillo, G., Faugeras, O.:
\newblock Flows of diffeomorphisms for multimodal image registration.
\newblock In: Biomedical Imaging, 2002. Proceedings. 2002 IEEE International
  Symposium on, IEEE (2002)  753--756

\bibitem{sundaramoorthi2007sobolev}
Sundaramoorthi, G., Yezzi, A., Mennucci, A.C.:
\newblock Sobolev active contours.
\newblock International Journal of Computer Vision \textbf{73}(3) (2007)
  345--366

\bibitem{calder2010image}
Calder, J., Mansouri, A., Yezzi, A.:
\newblock Image sharpening via sobolev gradient flows.
\newblock SIAM Journal on Imaging Sciences \textbf{3}(4) (2010)  981--1014

\bibitem{maaten2008visualizing}
Maaten, L.v.d., Hinton, G.:
\newblock Visualizing data using t-sne.
\newblock Journal of machine learning research \textbf{9}(Nov) (2008)
  2579--2605

\bibitem{erhan2009visualizing}
Erhan, D., Bengio, Y., Courville, A., Vincent, P.:
\newblock Visualizing higher-layer features of a deep network.
\newblock University of Montreal \textbf{1341}(3) (2009) ~1

\bibitem{simonyan2013deep}
Simonyan, K., Vedaldi, A., Zisserman, A.:
\newblock Deep inside convolutional networks: Visualising image classification
  models and saliency maps.
\newblock arXiv preprint arXiv:1312.6034 (2013)

\bibitem{zeiler2014visualizing}
Zeiler, M.D., Fergus, R.:
\newblock Visualizing and understanding convolutional networks.
\newblock In: European conference on computer vision, Springer (2014)  818--833

\bibitem{springenberg2014striving}
Springenberg, J.T., Dosovitskiy, A., Brox, T., Riedmiller, M.:
\newblock Striving for simplicity: The all convolutional net.
\newblock arXiv preprint arXiv:1412.6806 (2014)

\bibitem{nguyen2015deep}
Nguyen, A., Yosinski, J., Clune, J.:
\newblock Deep neural networks are easily fooled: High confidence predictions
  for unrecognizable images.
\newblock In: Proceedings of the IEEE Conference on Computer Vision and Pattern
  Recognition. (2015)  427--436

\bibitem{nguyen2016plug}
Nguyen, A., Yosinski, J., Bengio, Y., Dosovitskiy, A., Clune, J.:
\newblock Plug \& play generative networks: Conditional iterative generation of
  images in latent space.
\newblock arXiv preprint arXiv:1612.00005 (2016)

\bibitem{fong2017interpretable}
Fong, R.C., Vedaldi, A.:
\newblock Interpretable explanations of black boxes by meaningful perturbation.
\newblock arXiv preprint arXiv:1704.03296 (2017)

\bibitem{kindermans2017patternnet}
Kindermans, P.J., Sch{\"u}tt, K.T., Alber, M., M{\"u}ller, K.R., D{\"a}hne, S.:
\newblock Patternnet and patternlrp--improving the interpretability of neural
  networks.
\newblock arXiv preprint arXiv:1705.05598 (2017)

\bibitem{kindermans2017reliability}
Kindermans, P.J., Hooker, S., Adebayo, J., Alber, M., Sch{\"u}tt, K.T.,
  D{\"a}hne, S., Erhan, D., Kim, B.:
\newblock The (un) reliability of saliency methods.
\newblock arXiv preprint arXiv:1711.00867 (2017)

\bibitem{sundararajan2017axiomatic}
Sundararajan, M., Taly, A., Yan, Q.:
\newblock Axiomatic attribution for deep networks.
\newblock arXiv preprint arXiv:1703.01365 (2017)

\bibitem{grun2016taxonomy}
Gr{\"u}n, F., Rupprecht, C., Navab, N., Tombari, F.:
\newblock A taxonomy and library for visualizing learned features in
  convolutional neural networks.
\newblock arXiv preprint arXiv:1606.07757 (2016)

\bibitem{szegedy2013intriguing}
Szegedy, C., Zaremba, W., Sutskever, I., Bruna, J., Erhan, D., Goodfellow, I.,
  Fergus, R.:
\newblock Intriguing properties of neural networks.
\newblock arXiv preprint arXiv:1312.6199 (2013)

\bibitem{oygard2015glasses}
Øygard, A.M.:
\newblock Visualizing googlenet classes.
\newblock
  \url{https://www.auduno.com/2015/07/29/visualizing-googlenet-classes/} (2015)
  [Online; accessed March 13 2018].

\bibitem{mordvintsev2015inceptionism}
Mordvintsev, A., O.C.T.M.:
\newblock Inceptionism: Going deeper into neural networks.
\newblock
  \url{https://research.googleblog.com/2015/06/inceptionism-going-deeper-into-neural.html}
  (2015) [Online; accessed March 13 2018].

\bibitem{tyka2016art}
Tyka, M.:
\newblock Class visualization with bilateral filters.
\newblock
  \url{https://mtyka.github.io/deepdream/2016/02/05/bilateral-class-vis.html}
  (2016) [Online; accessed March 13 2018].

\bibitem{mordvintsev2016deepdream}
et~al., M.:
\newblock Deepdreaming with tensorflow.
\newblock
  \url{https://github.com/tensorflow/tensorflow/blob/master/tensorflow/examples/tutorials/deepdream/deepdream.ipynb}
  (2016) [Online; accessed March 13 2018].

\bibitem{dosovitskiy2015inverting}
Dosovitskiy, A., Brox, T.:
\newblock Inverting convolutional networks with convolutional networks.
\newblock CoRR abs/1506.02753 (2015)

\bibitem{nguyen2016synthesizing}
Nguyen, A., Dosovitskiy, A., Yosinski, J., Brox, T., Clune, J.:
\newblock Synthesizing the preferred inputs for neurons in neural networks via
  deep generator networks.
\newblock In: Advances in Neural Information Processing Systems. (2016)
  3387--3395

\bibitem{ulyanov2017practical}
Dmitry~Ulyanov, Andrea~Vedaldi, V.L.:
\newblock A practical test for univariate and multivariate normality.
\newblock Technical report, Skolkovo Institute of Science and Technology (2018)

\bibitem{christensen1996deformable}
Christensen, G.E., Rabbitt, R.D., Miller, M.I.:
\newblock Deformable templates using large deformation kinematics.
\newblock IEEE transactions on image processing \textbf{5}(10) (1996)
  1435--1447

\bibitem{fischer2004unified}
Fischer, B., Modersitzki, J.:
\newblock A unified approach to fast image registration and a new curvature
  based registration technique.
\newblock Linear Algebra and its applications \textbf{380} (2004)  107--124

\bibitem{rudin1992nonlinear}
Rudin, L.I., Osher, S., Fatemi, E.:
\newblock Nonlinear total variation based noise removal algorithms.
\newblock Physica D: nonlinear phenomena \textbf{60}(1-4) (1992)  259--268

\bibitem{chambolle2010introduction}
Chambolle, A., Caselles, V., Cremers, D., Novaga, M., Pock, T.:
\newblock An introduction to total variation for image analysis.
\newblock Theoretical foundations and numerical methods for sparse recovery
  \textbf{9}(263-340) (2010)  227

\bibitem{he2016deep}
He, K., Zhang, X., Ren, S., Sun, J.:
\newblock Deep residual learning for image recognition.
\newblock In: Proceedings of the IEEE conference on computer vision and pattern
  recognition. (2016)  770--778

\bibitem{huang2017densely}
Huang, G., Liu, Z., Weinberger, K.Q., van~der Maaten, L.:
\newblock Densely connected convolutional networks.
\newblock In: Proceedings of the IEEE conference on computer vision and pattern
  recognition. Volume~1. (2017) ~3

\bibitem{perona1990scale}
Perona, P., Malik, J.:
\newblock Scale-space and edge detection using anisotropic diffusion.
\newblock IEEE Transactions on pattern analysis and machine intelligence
  \textbf{12}(7) (1990)  629--639

\bibitem{kukavcka2017regularization}
Kuka{\v{c}}ka, J., Golkov, V., Cremers, D.:
\newblock Regularization for deep learning: A taxonomy.
\newblock arXiv preprint arXiv:1710.10686 (2017)

\end{thebibliography}
\end{document}